# Deep Air Learning: Interpolation, Prediction, and Feature Analysis of Fine-grained Air Quality

Zhongang Qi, Tianchun Wang, Guojie Song, Weisong Hu, Xi Li*, Zhongfei (Mark) Zhang

**Abstract**—The interpolation, prediction, and feature analysis of fine-gained air quality are three important topics in the area of urban air computing. The solutions to these topics can provide extremely useful information to support air pollution control, and consequently generate great societal and technical impacts. Most of the existing work solves the three problems separately by different models. In this paper, we propose a general and effective approach to solve the three problems in one model called the Deep Air Learning (DAL). The main idea of DAL lies in embedding feature selection and semi-supervised learning in different layers of the deep learning network. The proposed approach utilizes the information pertaining to the unlabeled spatio-temporal data to improve the performance of the interpolation and the prediction, and performs feature selection and association analysis to reveal the main relevant features to the variation of the air quality. We evaluate our approach with extensive experiments based on real data sources obtained in Beijing, China. Experiments show that DAL is superior to the peer models from the recent literature when solving the topics of interpolation, prediction, and feature analysis of fine-grained air quality.

✦

## 1 INTRODUCTION

THE interpolation, prediction, and feature analysis of fine-gained air quality are three important topics in the area of urban air computing. A good interpolation solves the problem that there are limited air-quality-monitor-stations whose distribution is uneven in a city; a precise prediction provides valuable information to protect humans from being damaged by air pollution; a reasonable feature analysis reveals the main relevant factors to the variation of air quality. In general, the solutions to these topics can extract extremely useful information to support air pollution control, and consequently generate great societal and technical impacts.

However, there exist several challenges for urban air computing as the related data have some special characteristics. First, since there are insufficient air-quality-monitor-stations in a city due to the high cost of building and maintaining such a station, it is expensive to obtain labeled

- Z. Qi is with the School of Electrical Engineering and Computer Science, Oregon State University, 1148 Kelley Engineering Center, Corvallis, OR 97331-5501, USA.
  E-mail: qiz@oregonstate.edu
- T. Wang is with the School of Information Systems, Singapore Management University, 178902, Singapore.
  E-mail: tcwang@smu.edu.sg
- G. Song is with the Key Laboratory of Machine Perception (Ministry of Education), Peking University, Beijing 100871, China.
  E-mail: gjsong@pku.edu.cn
- W. Hu is with the NEC Laboratories China, 11F Bldg. A, Innovation Plaza, Tsinghua Science Park Haidian District, Beijing 100084, China.
  E-mail: hu_weisong@nec.cn
- X. Li* (corresponding author) is with the College of Computer Science and Technology, Zhejiang University, No. 38, Zheda Road, Hangzhou 310027, China.
  E-mail: xilizju@zju.edu.cn
- Z. Zhang is with Computer Science Department, Watson School, State University of New York, Binghamton, NY 13902-6000, USA.
  E-mail: zzhang@binghamton.edu

training samples when dealing with fine-gained air quality. Second, the labeled data of the air-quality-monitor-stations are incomplete, and there exist lots of missing labels of the historical data in some time periods for some stations. The reason for the incomplete labels is related to the air quality monitor devices. In general, each station only has one monitor device which needs to be maintained at intervals, thus there will be no outputs for the station when the device is being maintained, recalibrated, or has other problems. Third, the kinds of urban air related data are various for the development of data acquisition technologies. However, there is not an universally accepted judgment to reveal the main causes of the occurrence and dissipation of air pollution, especially the pollution of $PM_{2.5}$. Hence, it is hard to know that what kinds of data are the main relevant features for interpolation and prediction, and the key factors for environment departments to prevent and control air pollution.

This paper is motivated to address all these challenges by utilizing the information contained in the unlabeled data and the spatio-temporal data, and performing feature selection and association analysis for the urban air related data. Though labeled data are difficult or expensive to obtain, large amounts of unlabeled examples can often be gathered cheaply. In general, unlabeled data can help in providing information to better exploit the geometric structure of the data. Moreover, most of the urban air related data contain both space and time information. In Figure 1, (a) and (b) show totally different observations in each place with a long time interval[1]; Each row in (c) shows continuous air quality changing observed in one place[2]; (d), (e), and (f) show observations at different spatial locations of Beijing at the same time[3], where Tian'anmen Square is close to Temple of Heaven, and far away from Beijing yanqing. Figure 1 shows an important statistical characteristic of spatio-temporal data, that is nearby (in space and time) observations tend to be more alike than those far apart, which can be utilized to complete the missing labels of the historical labeled data. Figure 2(a) shows the decision planes learned only from labeled data and learned from labeled and unlabeled data for the binary classification problem. Figure 2(b) shows the characteristic of the spatio-temporal data. Obviously, the information of the unlabeled data and the spatio-temporal data are valuable to guide the learning process and enhance the final performance.

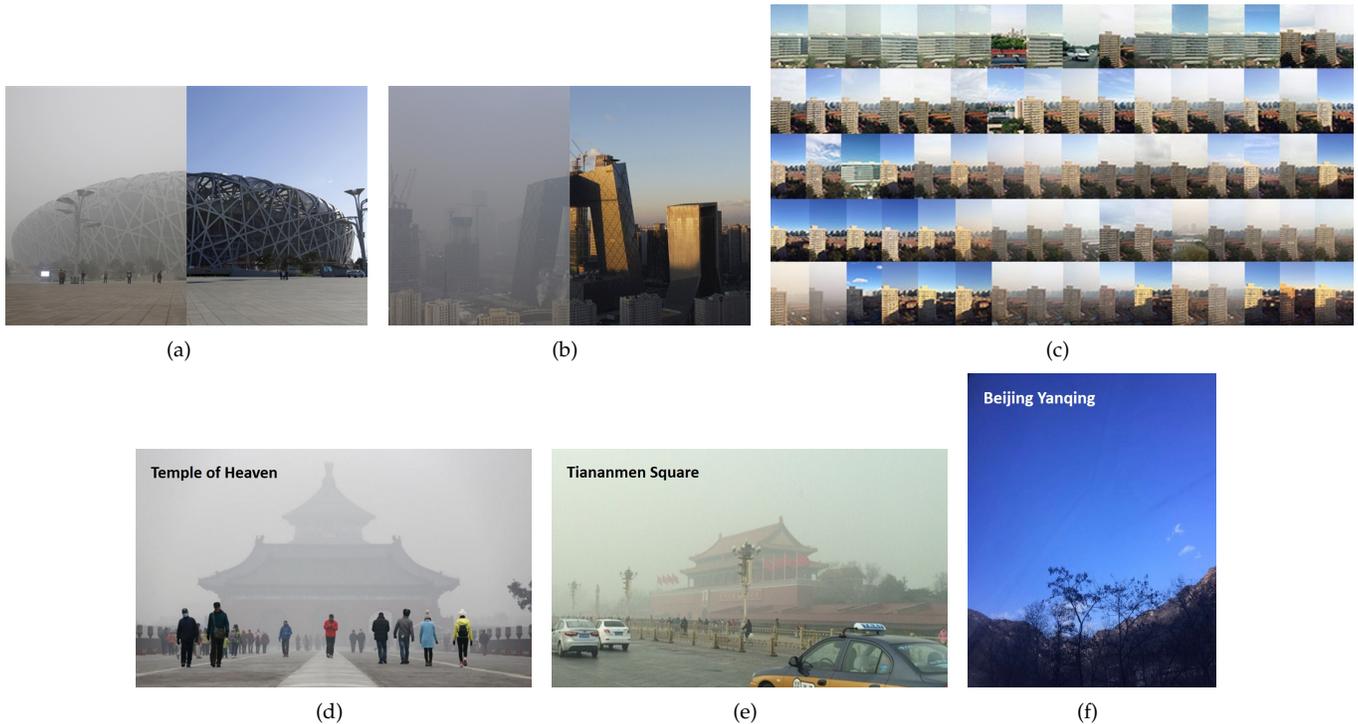

**Fig. 1: (a) and (b) show totally different observations with a long time interval in National Olympic Stadium (Bird Nest) and CCTV Headquarters, respectively; Each row in (c) shows continuous air quality changing of 16 days in 2015 at the same location of Beijing; (d), (e), and (f) are air quality comparisons among Temple of Heaven, Tian'anmen Square, and Beijing Yanqing at the same time in Beijing.**

Feature selection is the process of selecting a subset of relevant features for use in model construction, which reduces the dimension of the urban air related feature vector by ignoring the redundant or irrelevant data. Combining with feature selection, association analysis reveals the main relevant features to the variation of air quality. There exist several embedded methods incorporating feature selection into the regression based methods, such as Lasso [1] and Group Lasso [2]. However, utilizing the $L_1$-type penalty or the grouped $L_1$-type penalty to embed feature selection in the input layer of neural network is still challenging for the loss function of neural network is complex, nonlinear, and non-convex. In this paper, we propose a novel method which introduces an extra sparse layer to perform feature selection in neural network. The optimization of the pro-posed method is easy to solve and the performance of this method is well in finding the most relevant features in input layer to the final predictions.

Despite widespread adoption, neural networks based models remain black boxes for human. Understanding the reasons behind predictions is important for most applications, because it decides whether one can take actions based on the model's predictions, especially for air pollution's prevention and control. The proposed feature selection and analysis method reveals the importance of different input features to the predictions of the neural networks, thus has the ability to reveal some inner mechanism of the black-box deep models, which does not limit to air pollution prevention and control, but can also be applied to many other applications, such as medical diagnosis and terrorism detection. For example, one cannot take actions based on a model's predictions without knowing whether the black-box model is trustworthy or not, otherwise the consequences may be catastrophic.

Most of the existing work solve the problems of interpolation, prediction, and feature analysis of fine-grained air quality separately by different models. In this paper, we propose a general and effective approach to solve the three problems in one model called the Deep Air Learning (DAL), which addresses all the challenges that exist in the area of air computing simultaneously by a deep learning network. We claim that firstly, combining prediction with feature analysis is needed, because feature analysis helps to find out what are the main relevant features to the black-box model's predictions, and whether the prediction model is reasonable and trustworthy or not. Except for the chemical analysis, the data mining method can also provide useful information which helps the environment departments to understand the formation of air pollution, and thus to prevent and control the air pollution. Secondly, combining prediction with interpolation is convenient, because interpolation can also be considered as a prediction problem. Adding 1-dim to the output layer of the deep network can integrate interpolation with prediction, which almost does not increase the computation complexity of the prediction model. We first propose two novel methods to embed feature selection and spatiotemporal semi-supervised learning in the neural network, respectively. Then the framework of DAL is introduced and the details of association analysis are described. We evaluate our approach with extensive experiments based on real data sources obtained in Beijing, China. Experiments show that DAL is superior to the peer models from the recent literature when solving the topics of interpolation, prediction and feature analysis of fine-gained air quality.

---

3. photos are taken on December 25, 2015 in Beijing, from weather.com.cn and Chinanews.com

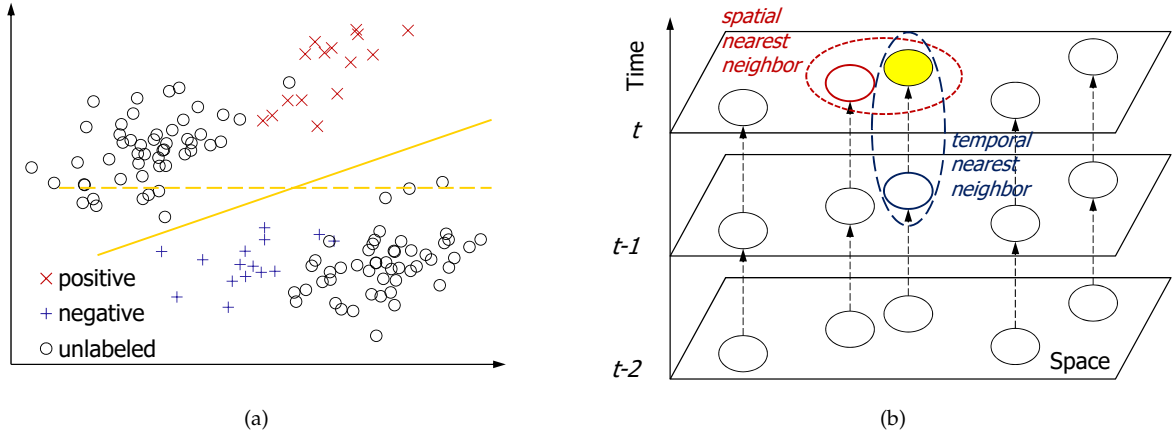

**Fig. 2: (a) The decision plane learned only from labeled data (the dashed line), and the decision plane learned from labeled and unlabeled data (the solid line); (b) The spatio-temporal data.**

Overall, our main contributions are as follows:

1) We develop a general and effective approach called DAL to unify the interpolation, prediction, feature selection and analysis of fine-grained air quality into one model.
2) We propose a novel method to perform feature selection in the input layer of neural network, whose optimization is easy to solve and performance is well in finding the most relevant features. The proposed feature selection and analysis method has the ability to reveal some inner mechanism of the black-box deep network models.
3) We utilize the characteristics of the spatio-temporal data, and the information contained in the unlabeled data, to achieve spatio-temporal semi-supervised learning in the neural network.
4) We demonstrate through extensive evaluations on real datasets that the DAL model performs well in comparison with the peer methods in the literature.

## 2 RELATED WORK

In this section, we discuss three main research categories related to our work, including air pollution control, semisupervised learning and feature selection.

**Air pollution control:** Various approaches have been proposed to apply data mining to the topics of interpolation, prediction, and feature analysis for air pollution control in the recent literature. For interpolation, Li et. al [3] investigates spatio-temporal interpolation methods for the application of air pollution assessment; Zheng et al. [4] infer the real-time and fine-grained air quality information throughout a city by a co-training-based approach; Hsieh et al. [5] propose an affinity graph framework to solve the problems of inferring real-time air quality of a given location and recommending the best locations where new monitor stations can be established. For prediction, Dong et al. [6] propose a novel framework based on hidden semi-Markov models (HSMMs) to predict high $PM_{2.5}$ concentration value; Thomas and Jacko [7] apply the regression and neural network models to forecasting expressway $PM_{2.5}$ concentration; Zheng et. al [8] build a real-time air quality forecasting system that uses data-driven models to predict fine-grained air quality over the following 48 hours. For feature analysis, Zhou et. al [9] propose a probabilistic dynamic causal model to uncover the dynamic temporal dependencies of $PM_{2.5}$; Singh et. al [10] utilize principal components analysis (PCA) to identify air pollution sources, and tree based ensemble learning models to predict the urban air quality in India. Different from most of the existing work which focus on one or two specific topics in the area of air pollution controlling, our work unify the interpolation, prediction, feature selection and analysis of fine-grained air quality into one model.

**Semi-supervised learning:** Semi-supervised learning is a class of supervised learning tasks and techniques that also make use of unlabeled data for training - typically a small amount of labeled data with a large amount of unlabeled data. Self-training [11] and co-training [12], [13] extract the most confidently classified examples from the unlabeled data, and add them into the labeled training set iteratively. Li et al. [14] explore the geometric structure of the marginal distribution of the whole data including the labeled and the unlabeled data through a specific family of parametric functions. Socher et al. [15] propose a recursive autoencoder trained on both labeled and unlabeled texts to predict sentiment distribution. Weston et al. [16] show how one can improve supervised learning for deep architectures if one jointly learns an embedding task using unlabeled data. For spatio-temporal data, the nearby (in space and time) observations tend to be more alike than those far apart [17]. In our work, this characteristic is utilized in semi-supervised learning to better exploit the geometric structure of the spatio-temporal data, and to achieve the purpose of interpolation.

**Feature selection:** In many real-world applications, feature selection techniques have become an apparent need. Saeys et al. [18] give a review of feature selection techniques in bioinformatics. Setiono and Liu [19] propose a three-layer feedforward neural network trained on the whole set of attributes and removing the irrelevant features one by one. Also, there exist several embedded methods incorporating feature selection. The $L_1$-type penalty of the lasso [1], which was originally proposed for linear regression models, has also been applied to other models as for example Cox regression [20], logistic regression [21], [22], and multinomial logistic regression [23]. Ng [24] compares the $L_1$ regularization and the $L_2$ regularization on logistic regression for preventing overfitting. Group lasso [2], [25], [26] introduces a penalty which can be viewed as an intermediate between the $L_1$- and $L_2$-type penalty to select grouped features instead of individual variables. The main drawback of the $L_1$-type penalty is its non-smooth characteristic thus leading to complex optimization effort. In particular, the LARS algorithm [27] allows to find the entire regularization path at the cost of a single matrix inversion. Proximal-gradient methods [28] use an approximate proximity operator for taking advantage of the structure of the non-smooth optimization problem. However, these optimization methods are difficult to be performed in neural network framework, for the loss function of neural network is complex, nonlinear, and non-convex. Different from the work of removing the irrelevant features one by one in neural network [19], and the work of adding the $L_1$-type penalty, our work introduces an extra sparse layer to remove all the redundant or irrelevant features during one training process, whose optimization is easy to solve.

## 3 MODEL FORMULATION

The main idea of this work is based on embedding feature selection and spatio-temporal semi-supervised learning simultaneously in different layers of the deep learning network. Considering the topics of interpolation and prediction both as the classification problem with different outputs, we use a general multiple-output classifier to solve the two topics. In this paper we propose a novel deep learning network as the multiple-output classifier which utilizes the information pertaining to the unlabeled spatio-temporal data not only to achieve the purpose of interpolation, but also to improve the performance of the prediction. Further, the main relevant features to the variation of the air quality can also be revealed by embedding feature selection and performing association analysis in the proposed framework. We begin with describing the data representations and symbols used in this work, then introduce the method to embed feature selection in the neural network. The method of how to utilize the information of the unlabeled spatio-temporal data is described in Section 3.3. We introduce the general framework of DAL and give the details of association analysis in Section 3.4. Finally we describe the learning algorithm for DAL in Section 3.5.

### 3.1 Data Representations and Symbols

We denote a training dataset as $\mathcal{I}$ which contains $m$ labeled instances $\{(\mathbf{x}^{(i)}, \mathbf{y}^{(i)})\}_{i=1}^{m}$ and $u$ unlabeled instances $\{\mathbf{x}^{(i)}\}_{i=1}^{u}$. $\mathbf{x}^{(i)} \in \mathbb{R}^K$ is the feature descriptor for instance $i \in \mathcal{I}$, and $\mathbf{y}^{(i)} \in \mathbb{R}^O$ is the corresponding label for the labeled training data. $x_p^{(i)}$ and $y_p^{(i)}$ are the $p$-th element in vector $\mathbf{x}^{(i)}$ and vector $\mathbf{y}^{(i)}$, respectively. Given a neural network, let $n_l$ denote the number of the layers. Hence, layer 1 is the input layer and layer $n_l$ is the output layer. The neural network has parameters $(W_{ji}^{(l)}, b_{ji}^{(l)})$, where $W_{ji}^{(l)}$ is the weight associated with the connection between unit $i$ in layer $l$ and unit $j$ in layer $l + 1$, $b_j^{(l)}$ is the bias associated with unit $j$ in layer $l + 1$. Note that the bias units do not have inputs or connections going into them for they always output the value $+1$. We also let $s_l$ denote the number of nodes in layer $l$ (not counting the bias unit). $W^{(l)}$ is considered as a $s_{l+1} \times s_l$ matrix the element of which is $W_{ji}^{(l)}$. $W_{j.}^{(l)}$ is the $j$th row in $W^{(l)}$. $a_j^{(l)}$ is denoted as the activation of node $j$ in the layer $l$, $z_j^{(l)}$ is denoted as the total weighted sum of inputs to node $j$ in layer $l$, and $f$ is denoted as the activation function. Thus, $a_j^{(l)} = f(z_j^{(l)})$. $h_{W,b}(\mathbf{x}^{(i)})$ is the output of our hypothesis on input $\mathbf{x}^{(i)}$ using parameters $W$, $b$, which is a vector of the same dimension as the target value $\mathbf{y}^{(i)}$. $h_{W,b}(\mathbf{x}^{(i)})_q$ is the $q$-th element in vector $h_{W,b}(\mathbf{x}^{(i)})$.

### 3.2 Feature Selection in Neural Network

Most of the existing neural network based methods focus on imposing sparsity constraints on the hidden units instead of on the input units to discover the interesting structure in the data. In this paper, we claim that feature selection for the input units is also important in the area of neural network even if some deep learning network based methods have the ability of feature learning, especially when there are many irrelevant input features and we intend to learn the correlation between the target and the input features. Embedding feature selection in neural network encourages the parameter vector for the input layer to be sparse. Hence, many irrelevant features are ignored and the main relevant features to the target are revealed explicitly for association analysis.

Because of the nonlinearity and the complexity, neural networks based models remain black boxes for human. In most cases, we observe that deep networks models achieve good performance in many applications, without knowing how these models work and whether their predictions are reasonable and trustworthy or not. The proposed feature selection and analysis method has the ability to reveal some inner mechanism of the black box by discovering the importance of each input feature to the model's final predictions, which provides useful information not only to air pollution's prevention and control, but also to many other applications where knowing how the model works is important.

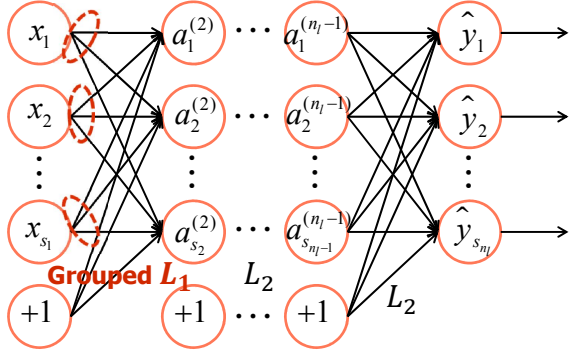

**Fig. 3: The graph of the neural network with grouped $L_1$ in the input layer.**

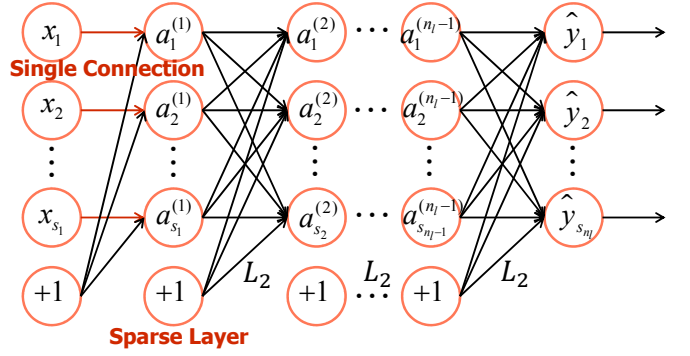

**Fig. 4: The graph of the neural network with feature selection.**

$L_1$ regularization uses a penalty term which encourages the sum of the absolute values of the parameters to be small. In many models $L_1$ regularization causes many parameters to equal zero, which makes it a natural candidate in feature selection. Grouped $L_1$ regularization selects the whole factors of a group instead of the individual dummy variables. It has the attractive property when embedding in neural network that it truly does the feature selection instead of parameter selection by treating all the parameters of each feature as a natural group. Figure 3 shows the graph of the neural network with grouped $L_1$ in the input layer. The cost function of the network is presented as follows:

$$J(W, b) = \frac{1}{m} \sum_{i=1}^{m} L(W, b; \mathbf{x}^{(i)}, \mathbf{y}^{(i)}) + \lambda_1 ||W^{(1)}||_{2,1}$$
$$+ \frac{\lambda_2}{2} \sum_{l=2}^{n_l-1} ||W^{(l)}||_2^2 \tag{1}$$

The first term in the definition of $J(W, b)$ is the loss function over the training set of $m$ examples. The second term is the grouped $L_1$ regularization for the input layer to perform feature selection. The third term is the square of the $L_2$ regularization for the other layers to avoid over-fitting. However, the second term in optimization (1) is nonsmooth, and the first term is a complex nonlinear and non-convex function for neural network. Hence, it is difficult to utilize the existing methods, i.e. LARS [27], the proximal method [28], the coordinate descent method [29], to solve the optimization.

In this paper, we propose another novel method which introduces an extra sparse layer between the input layer and the second layer in neural network to perform feature selection. The number of the nodes in the extra sparse layer is the same as that in the input layer, and there only exists a single connection between the node in the input layer and the node in the extra sparse layer, as Figure 4 shows. Here the input layer becomes layer 0, and the extra sparse layer becomes layer 1. We denote the weight matrix between the input layer and the extra sparse layer as $W^{(0)}$, which is a diagonal matrix. $W^{(0)} = diag(W_{11}^{(0)}, W_{22}^{(0)}, ..., W_{s_1 s_1}^{(0)})$. For initialization, $W_{11}^{(0)} = W_{22}^{(0)} = ... = W_{s_1 s_1}^{(0)}$. Different from the method introduced by sparse autoencoder [30], we add the sparse constraints on the extra sparse layer to perform feature selection on the input layer. The cost function of the network showed in Figure 4 is presented as follows:

$$J(W, b) = \frac{1}{m} \sum_{i=1}^{m} L(W, b; \mathbf{x}^{(i)}, \mathbf{y}^{(i)}) + \lambda_1 \sum_{j=1}^{s_1} \mathrm{KL}(\rho || \hat{\rho}_j)$$
$$+ \frac{\lambda_2}{2} \sum_{l=0}^{n_l-1} ||W^{(l)}||_2^2 \tag{2}$$

where $\mathrm{KL}(\rho || \hat{\rho}_j)$ is the Kullback-Leibler Divergence between $\rho$ and $\hat{\rho}_j$, $\hat{\rho}_j$ is an average result of the activation $a_j^{(1)}$ for node $j$ in the extra sparse layer over the whole training set, and $\rho$ is the sparsity parameter.

$$\mathrm{KL}(\rho || \hat{\rho}_j) = \rho \log \frac{\rho}{\hat{\rho}_j} + (1 - \rho) \log \frac{1 - \rho}{1 - \hat{\rho}_j}$$
$$\hat{\rho}_j = \frac{1}{m} \sum_{i=1}^{m} a_j^{(1)}(\mathbf{x}^{(i)})$$
$$a_j^{(1)}(\mathbf{x}^{(i)}) = f(\langle W_{j.}^{(0)}, \mathbf{x}^{(i)} \rangle + b_j^{(0)}) = f(W_{jj}^{(0)} x_j^{(i)} + b_j^{(0)}) \tag{3}$$

The sparsity parameter $\rho$ is typically a small value close to zero. The KL Divergence constraint $\mathrm{KL}(\rho || \hat{\rho}_j)$ would make the average activation of each node $j$ in the extra sparse layer to be close to $\rho$. The activation of node $j$ in the extra sparse layer only has a single connection with node $j$ in the input layer. To satisfy this constraint, some of the activations of the nodes in the sparse layer must be near zero. Hence, the effect of the corresponding features in the input layer is removed from the network respectively to achieve the purpose of feature selection. The irrelevant features can be ignored and the main relevant features to the target can be revealed during the training of the network. Compared with optimization (1), the optimization (2) is simple to be solved with a small change to the Back Propagation algorithm [31].

### 3.3 Spatio-temporal Semi-supervised Learning in Neural Network

A limitation of supervised learning is that it requires a set of labeled examples which are often difficult or expensive to obtain. However, large amounts of unlabeled examples can often be gathered cheaply. The information contained in plentiful unlabeled examples can be utilized to better exploit the geometric structure of the data, especially for the spatio-temporal data. An important statistical characteristic of spatio-temporal data is that nearby (in space and time) observations tend to be more alike than those far apart. Based on this characteristic, we propose a novel method which embeds spatio-temporal semi-supervised learning in the output layer of the neural network by minimizing the following loss function between the nearby observations over the labeled and unlabeled training set.

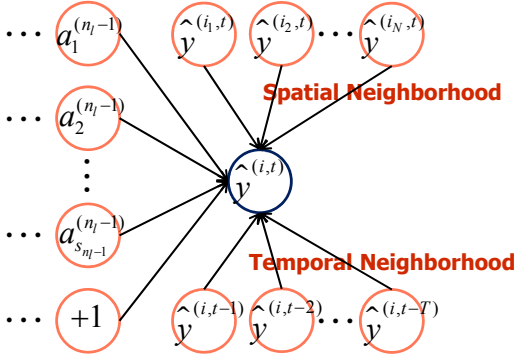

**Fig. 5: The graph of the spatio-temporal semi-supervised neural network.**

$$\frac{1}{m+u} \sum_{i=1}^{m+u} \sum_{j \in \mathcal{N}_i} e^{-\left(d_s(\mathbf{x}^{(i)}, \mathbf{x}^{(j)}) + \alpha \cdot d_t(\mathbf{x}^{(i)}, \mathbf{x}^{(j)})\right)} \bar{L}(W, b; \mathbf{x}^{(i)}, \mathbf{x}^{(j)})$$

(4)

where $\mathcal{N}_i$ is the spatio-temporal neighborhood of instance $i$, $d_s$ is the spatial distance measure, $d_t$ is the temporal distance measure, and $\alpha$ is a parameter. Selecting the quadratic loss as the loss function, we obtain: $\bar{L}(W, b; \mathbf{x}^{(i)}, \mathbf{x}^{(j)}) = \frac{1}{2}||h_{W,b}(\mathbf{x}^{(i)}) - h_{W,b}(\mathbf{x}^{(j)})||^2$.

To simplify (4), we only consider the spatial loss between the observations at the same time and the temporal loss between the observations at the same location, and obtain the following loss function:

$$\frac{1}{m+u} \sum_{i=1}^{m+u} \sum_{j \in \mathcal{N}_i^{(spa)}} C_{ij}^{(s)} \bar{L}(W, b; \mathbf{x}^{(i)}, \mathbf{x}^{(j)})$$

$$+ \frac{\alpha}{m+u} \sum_{i=1}^{m+u} \sum_{j \in \mathcal{N}_i^{(tem)}} C_{ij}^{t} \bar{L}(W, b; \mathbf{x}^{(i)}, \mathbf{x}^{(j)})$$

(5)

where $\mathcal{N}_i^{(spa)}$ is the spatial neighborhood of instance $i$ at the same time, $\mathcal{N}_i^{(tem)}$ is the temporal neighborhood of instance $i$ at the same location, $C_{ij}^{(s)} = e^{-d_s(\mathbf{x}^{(i)}, \mathbf{x}^{(j)})}$, and $C_{ij}^{(t)} = e^{-d_t(\mathbf{x}^{(i)}, \mathbf{x}^{(j)})}$.

Figure 5 shows the graph of the spatio-temporal semi-supervised neural network. In Figure 5 we choose $N$ nearest neighbors $i_1, i_2, \ldots, i_N$ as the spatial neighborhood for instance $i$ at time $t$, and select the previous $T$ times $t-1, t-1, \ldots, t-T$ of instance $i$ as the temporal neighborhood for instance $i$. Neural network is a general classifier which can have multiple outputs. For the purpose of representation clarity, we only draw one output node in Figure 5. The spatio-temporal semi-supervised learning is utilized in our work not only to achieve the purpose of interpolation, but also to improve the performance of prediction. When $t$ is the current time, $\hat{y}^{(i,t)}$ is the predicted label for interpolation; when $t$ is a future time, $\hat{y}^{(i,t)}$ is the predicted label for prediction.

### 3.4 Deep Air Learning

By embedding feature selection and spatio-temporal semi-supervised learning in the input layer and the output layer of the deep learning network respectively, we propose a general and effective approach called Deep Air Learning (DAL). The graph of DAL is shown in Figure 6. There exist a large amount of unlabeled data both in spatial dimension and temporal dimension, which can be utilized to pretrain the weights of the deep model.

In the input layer of DAL, we embed feature selection in an Autoencoder [32] over the whole training set with labeled and unlabeled data to pretrain the weights. The proposed model is called Autoencoder with Feature Selection (AEFS), and its cost function is presented as follows:

$$J(W, b) = \frac{1}{m+u} \sum_{i=1}^{m+u} L(W, b; \mathbf{x}^{(i)}, \mathbf{x}^{(i)}) + \lambda_1 \sum_{j=1}^{s_1} \text{KL}(\rho||\hat{\rho}_j)$$

$$+ \frac{\lambda_2}{2} \sum_{l=0}^{2} ||W^{(l)}||_2^2$$

(6)

In the middle layers of DAL, we use an Autoencoder (AE) directly over the whole training set with labeled and unlabeled data to pretrain the weights of these layers. The cost function of AE is presented as follows:

$$J(W, b) = \frac{1}{m+u} \sum_{i=1}^{m+u} L(W, b; \mathbf{x}^{(i)}, \mathbf{x}^{(i)}) + \frac{\lambda}{2} \sum_{l=1}^{2} ||W^{(l)}||_2^2$$

(7)

In the output layer of DAL, we embed spatio-temporal semi-supervised learning in regression to pretrain the weights. The proposed model is called Spatio-temporal

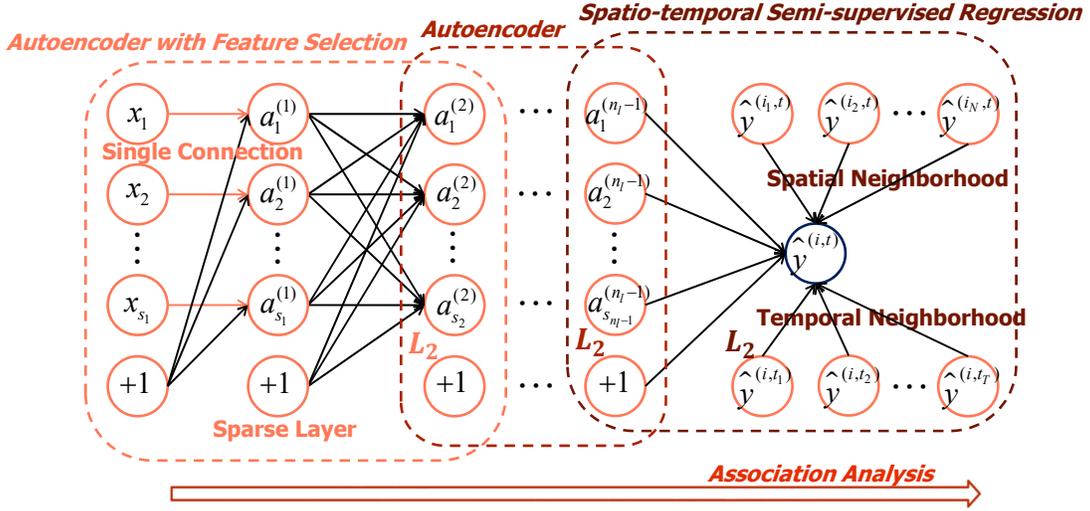

**Fig. 6: The graph of Deep Air Learning (DAL).**

Semi-supervised Regression (STSR), and its cost function is presented as follows:

$$J(W, b) = \frac{1}{m} \sum_{i=1}^{m} L(W, b; \mathbf{x}^{(i)}, \mathbf{y}^{(i)}) + \frac{\lambda}{2} ||W||_2^2$$

$$+ \frac{\alpha}{m+u} \sum_{i=1}^{m+u} \sum_{j \in \mathcal{N}_i^{(spa)}} C_{ij}^{(s)} \bar{L}(W, b; \mathbf{x}^{(i)}, \mathbf{x}^{(j)})$$

$$+ \frac{\beta}{m+u} \sum_{i=1}^{m+u} \sum_{j \in \mathcal{N}_i^{(tem)}} C_{ij}^{(t)} \bar{L}(W, b; \mathbf{x}^{(i)}, \mathbf{x}^{(j)}) \quad (8)$$

Finally, we obtain the overall cost function of DAL:

$$J(W, b) = \frac{1}{m} \sum_{i=1}^{m} L(W, b; \mathbf{x}^{(i)}, \mathbf{y}^{(i)})$$

$$+ \lambda_1 \sum_{j=1}^{s_1} \text{KL}(\rho || \hat{\rho}_j) + \frac{\lambda_2}{2} \sum_{l=0}^{n_1-1} ||\mathbf{W}^{(l)}||_2^2$$

$$+ \frac{\alpha}{m+u} \sum_{i=1}^{m+u} \sum_{j \in \mathcal{N}_i^{(spa)}} C_{ij}^{(s)} \bar{L}(W, b; \mathbf{x}^{(i)}, \mathbf{x}^{(j)})$$

$$+ \frac{\beta}{m+u} \sum_{i=1}^{m+u} \sum_{j \in \mathcal{N}_i^{(tem)}} C_{ij}^{(t)} \bar{L}(W, b; \mathbf{x}^{(i)}, \mathbf{x}^{(j)}) \quad (9)$$

We use (9) to perform fine tuning. The pretrained weights obtained from all the labeled and unlabeled data are utilized as the prior weights for the fine-tuning process, which is helpful to improve the prediction accuracy. Fine tuning treats all the layers of DAL as a single model, and consequently, the weights of all the layers in DAL can be improved in one iteration. The algorithm of the DAL approach is described in Algorithm 1.

We perform association analysis by normalizing each feature of the input before the training and evaluating the effect of each feature for the target after the training. The evaluation method is described in Algorithm 2. When $eff^k > 0$, feature $k$ is a relevant factor to the target; when $eff^k = 0$, feature $k$ is an irrelevant factor; when $eff^k < 0$,

---

**Algorithm 2:** Association Analysis in DAL

1 **Compute** the mean loss $l_s$ over $m$ labeled training instances and $n$ testing instances using the trained weights of DAL: $l_s = \frac{1}{m+n} \sum_{i=1}^{m+n} L(W, b; \mathbf{x}^{(i)}, \mathbf{y}^{(i)})$
2 **foreach** *feature* $k$, $1 \leq k \leq K$ **do**
3    **foreach** $i$, $1 \leq i \leq m+n$ **do**
4      **Let** $\hat{\mathbf{x}}^{(i)} = \mathbf{x}^{(i)}$, and **Set** $\hat{x}_k^{(i)} = 0$
5    **Compute** the mean loss $l_s^k$:
     $l_s^k = \frac{1}{m+n} \sum_{i=1}^{m+n} L(W, b; \hat{\mathbf{x}}^{(i)}, \mathbf{y}^{(i)})$
6    **Compute** the effect of feature $k$ for the target:
     $eff^k = l_s^k - l_s$

---

feature $k$ is a disturbance. After feature selection performed by DAL, some of the activations of the nodes in the sparse layer are near zero, thus the corresponding features which are irrelevant factors and disturbances in the input layer is removed from the network respectively. It is obvious that the larger the effect $eff$ is, the more relevant the feature is to the target when $eff > 0$.

Combining with feature selection, association analysis reveals the main relevant features to the variation of the target, which is helpful for us to understand how the blackbox prediction model works and whether its predictions are reasonable and trustworthy or not in general. The results of feature selection and analysis from a reasonable and trustworthy prediction model provide some insights for the formation of air pollution, which demonstrates the effectiveness of the data mining method in the application of air pollution prevention and control. Further, by embedding spatio-temporal semi-supervised learning in DAL, the purpose of interpolation is achieved. The output $\hat{y}^{(i,t)}$ is the predicted label for interpolation when $t$ is the current time, and is the predicted label for prediction when $t$ is a future time.

### 3.5 Learning Algorithm

Our goal is to minimize the cost function $J(W, b)$ as a function of $W$ and $b$. In this section, we describe the back-propagation learning algorithm for DAL, which gives an efficient way to solve the optimization of neural network with the extra sparse layer and the spatio-temporal semi-supervised constraints. Selecting the quadratic loss as the loss function, we obtain the overall cost function of DAL in (9) as follows:

$$
\begin{aligned}
J(W,b) &= \frac{1}{m}\sum_{i=1}^{m}\frac{1}{2}||h_{W,b}(\mathbf{x}^{(i)}) - \mathbf{y}^{(i)}||^2 \\
&+ \lambda_1\sum_{j=1}^{s_1}\mathrm{KL}(\rho||\hat{\rho}_j) + \frac{\lambda_2}{2}\sum_{l=0}^{n_l-1}||\mathrm{W}^{(l)}||_2^2 \\
&+ \frac{\alpha}{m+u}\sum_{i=1}^{m+u}\sum_{j\in\mathcal{N}_i^{(spa)}}\frac{C_{ij}^{(s)}}{2}||h_{W,b}(\mathbf{x}^{(i)}) - h_{W,b}(\mathbf{x}^{(j)})||^2 \\
&+ \frac{\beta}{m+u}\sum_{i=1}^{m+u}\sum_{j\in\mathcal{N}_i^{(tem)}}\frac{C_{ij}^{(t)}}{2}||h_{W,b}(\mathbf{x}^{(i)}) - h_{W,b}(\mathbf{x}^{(j)})||^2
\end{aligned}
\tag{10}
$$

In detail, here is the learning algorithm for DAL:

1. Perform a feedforward pass, computing the activations for layer 1 (the extra sparse layer), layer 2, layer 3, and so on up to the output layer $n_l$.

2. For each output node $q$ in layer $n_l$, set

$$
\delta_{qp}^{(n_l)(i)} = \frac{\partial h_{W,b}(\mathbf{x}^{(i)})_q}{\partial z_p^{(n_l)(i)}} = \begin{cases} f'(z_p^{(n_l)(i)}), & \text{if } q = p \\ 0, & \text{if } q \neq p \end{cases}
\tag{11}
$$

3. For layer $l = n_l - 1, n_l - 2, n_l - 3, ..., 2, 1$
For each node $p$ in layer $l$, set

$$
\delta_{qp}^{(l)(i)} = \frac{\partial h_{W,b}(\mathbf{x}^{(i)})_q}{\partial z_p^{(l)(i)}} = \left(\sum_{k=1}^{s_{l+1}}\delta_{qk}^{(l+1)(i)}W_{kp}^{(l)}\right)f'(z_p^{(l)(i)})
\tag{12}
$$

4. Compute the desired partial derivatives for layer $l = n_l, n_l - 1, n_l - 2, ..., 3, 2$, which are given as:

$$
\begin{aligned}
&\frac{\partial J(W,b)}{\partial W_{rp}^{(l-1)}} = \\
&\frac{1}{m}\sum_{i=1}^{m}\sum_{q=1}^{s_{n_l}}\left(h_{W,b}(\mathbf{x}^{(i)})_q - y_q^{(i)}\right)\delta_{qp}^{(l)(i)}a_p^{(l-1)(i)} + \lambda_2 W_{rp}^{(l-1)} \\
&+ \frac{\alpha}{m+u}\sum_{i=1}^{m+u}\sum_{j\in\mathcal{N}_i^{(spa)}}C_{ij}^{(s)}\sum_{q=1}^{s_{n_l}}\left(h_{W,b}(\mathbf{x}^{(i)})_q - h_{W,b}(\mathbf{x}^{(j)})_q\right) \\
&\cdot\left(\delta_{qp}^{(l)(i)}a_p^{(l-1)(i)} - \delta_{qp}^{(l)(j)}a_p^{(l-1)(j)}\right) \\
&+ \frac{\beta}{m+u}\sum_{i=1}^{m+u}\sum_{j\in\mathcal{N}_i^{(tem)}}C_{ij}^{(t)}\sum_{q=1}^{s_{n_l}}\left(h_{W,b}(\mathbf{x}^{(i)})_q - h_{W,b}(\mathbf{x}^{(j)})_q\right) \\
&\cdot\left(\delta_{qp}^{(l)(i)}a_p^{(l-1)(i)} - \delta_{qp}^{(l)(j)}a_p^{(l-1)(j)}\right)
\end{aligned}
\tag{13}
$$

The derivation process of formula $\frac{\partial J(W,b)}{\partial b^{(l-1)}}$ is similar as that of formula $\frac{\partial J(W,b)}{\partial W_{rp}^{(l-1)}}$.

5. Compute the desired partial derivatives for layer 1 (the extra sparse layer), which are given as:

$$
\begin{aligned}
&\frac{\partial J(W,b)}{\partial W_{pp}^{(0)}} = \frac{1}{m}\sum_{i=1}^{m}\sum_{q=1}^{s_{n_l}}\left(h_{W,b}(\mathbf{x}^{(i)})_q - y_q^{(i)}\right)\delta_{qp}^{(1)(i)}x_p^{(i)} \\
&+ \frac{\lambda_1}{m}\left(-\frac{\rho}{\hat{\rho}_p} + \frac{1-\rho}{1-\hat{\rho}_p}\right)\sum_{i=1}^{m}f'(z_p^{(1)(i)})x_p^{(i)} + \lambda_2 W_{pp}^{(0)} \\
&+ \frac{\alpha}{m+u}\sum_{i=1}^{m+u}\sum_{j\in\mathcal{N}_i^{(spa)}}C_{ij}^{(s)}\sum_{q=1}^{s_{n_l}} \\
&\left(h_{W,b}(\mathbf{x}^{(i)})_q - h_{W,b}(\mathbf{x}^{(j)})_q\right)\left(\delta_{qp}^{(1)(i)}x_p^{(i)} - \delta_{qp}^{(1)(j)}x_p^{(j)}\right) \\
&+ \frac{\beta}{m+u}\sum_{i=1}^{m+u}\sum_{j\in\mathcal{N}_i^{(tem)}}C_{ij}^{(t)}\sum_{q=1}^{s_{n_l}} \\
&\left(h_{W,b}(\mathbf{x}^{(i)})_q - h_{W,b}(\mathbf{x}^{(j)})_q\right)\left(\delta_{qp}^{(1)(i)}x_p^{(i)} - \delta_{qp}^{(1)(j)}x_p^{(j)}\right)
\end{aligned}
\tag{14}
$$

The derivation process of formula $\frac{\partial J(W,b)}{\partial b^{(0)}}$ is similar as that of formula $\frac{\partial J(W,b)}{\partial W_{pp}^{(0)}}$. Since $W^{(0)}$ is a diagonal matrix, we hold $W_{rp}^{(0)} = 0,\ r \neq p$ in the learning algorithm.

## 4 EXPERIMENTS

### 4.1 Data and Parameter Setting

In the evaluation we apply our model to the topics of interpolation, prediction, and feature analysis of fine-grained air quality based on real data sources obtained in Beijing, China. The real data sources used in the experiments contains the following datasets:

1) Meteorological data: We collect fine-gained (3km×3km) meteorological data, consisting of temperature, humidity, barometric pressure, precipitation, wind direction, and wind strength every hour, generated from a classic numerical weather prediction model we build in our labs.

2) Air quality data: We collect real valued concentration of six kinds of air pollutants, consisting of $PM_{2.5}$, $PM_{10}$, $SO_2$, $NO_2$, $CO$, and $O_3$ every hour, reported by 35 ground-based air quality monitor stations in Beijing.

We divide the Beijing city (within the Fifth Ring Road) into disjointed grids (e.g., 1km×1km in the experiments) assuming that each grid is an instance and the air quality in a grid is uniform. The instance is labeled if there is an air quality monitor station located in the grid; otherwise, the instance is unlabeled. In the experiments, 35 grids are labeled, and 3420 grids are unlabeled. Assuming that the current time is $t$, the targets are to interpolate the current $PM_{2.5}$'s concentration for unlabeled data, and to predict the $PM_{2.5}$'s concentration of the following $T_f$ hours for both the labeled and unlabeled data. Hence, the number of the output nodes for DAL is $T_f + 1$. For a given instance $i$, the input vector $\mathbf{x}^{(i)}$ contains the following features:

1) The meteorological data for the grid of instance $i$, from time $t - T_{p_1}$ to time $t + T_f$;

2) The air quality data of the nearest $N$ air quality monitor stations for the grid of instance $i$, from time $t - T_{p_2}$ to time $t$.

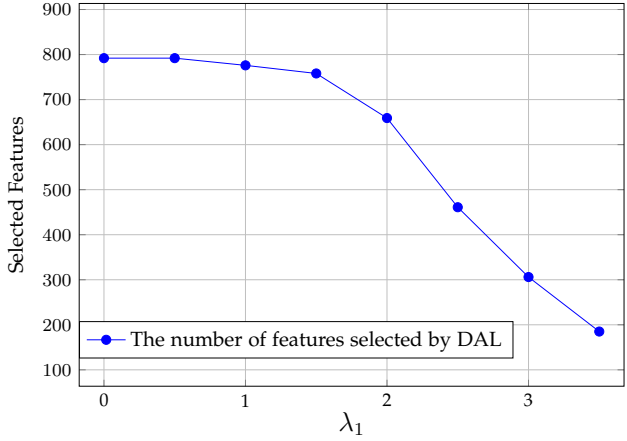

(a) The number of features selected by DAL

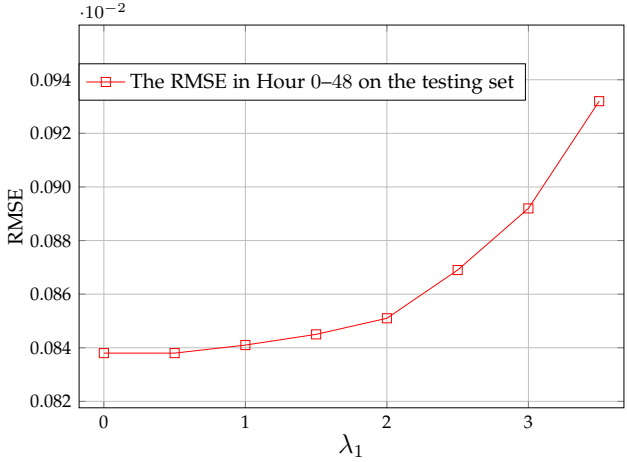

(b) The RMSE in Hour 0–48 on the testing set

Fig. 7: **The number of features selected by DAL and the corresponding RMSE on the testing set.**

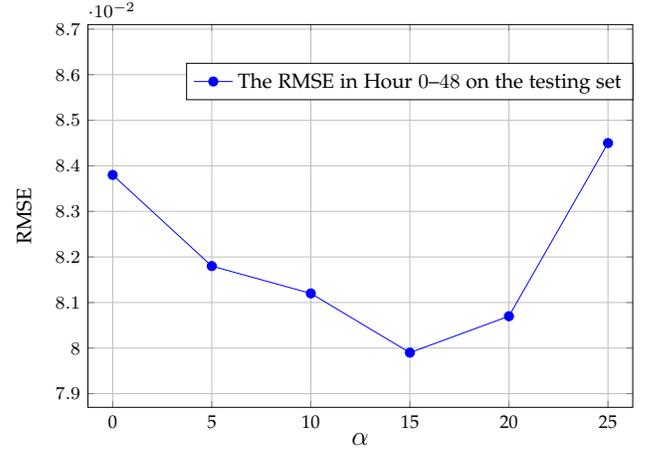

(a) The RMSE on the testing set as a function of $\alpha$.

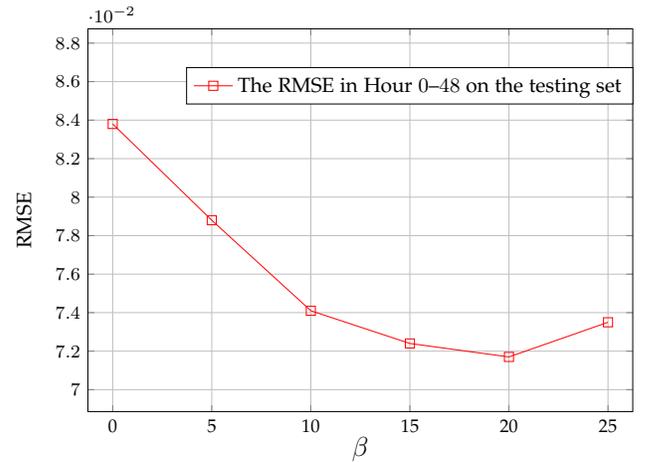

(b) The RMSE on the testing set as a function of $\beta$.

Fig. 8: **The RMSE on the testing set as a function of $\alpha$, and a function of $\beta$, respectively.**

In the experiments, we use real data sources of 60 days collected from Beijing in November and December, 2015. The $PM_{2.5}$'s concentration is our prediction target. We set $T_{p_1} = 11$, $T_{p_2} = 23$, $N = 3$, and $T_f = 48$, thus the dimensionality of the input vector is $6 \cdot (T_f + T_{p_1} + 1) + 6 \cdot N \cdot (T_{p_2} + 1) = 792$, and the dimensionality of the output vector is 49 (Hour 0–48). Since the number of the unlabeled grids is very large, every 7 grids we select a grid as the unlabeled instance. Hence, the number of the unlabeled grids is reduced to 489. We repeat 10 times experiments and report the averages of the results. For each experiment, we select 400 hours of data as the training set, and 300 hours of data as the testing set. The number of labeled and unlabeled training instances amount to $(35 + 489) \times 400 = 209600$, thus constructing a large-scale data set. Different training sets and testing sets are selected from the real data sources of 60 days for each of the 10 times experiments.

We compare our model DAL with several closely related methods, including Logistic Regression[4], $L_1$-regularized

Logistic Regression[5] [33], $L_1$-regularized Logistic Regression with the same spatio-temporal constraints (ST), Neural Network, Autoencoder, Laplacian Regression (LapReg) [34], Autoregressive Integrated Moving Average (ARIMA) [35], and Recurrent Neural Network (RNN) [36] in the experiments. The root-mean-square error (RMSE) is used to evaluate the performances of the methods. For simplicity, we set the numbers of the layers for Neural Network and Autoencoder both to 4, and the number of the layers for DAL to 5, since there is an extra sparse layer in DAL. The activation function for the neural network based methods is the sigmoid function, and the learning rate is 0.5. The size of the neighborhood for each instance $i$ is defined as the count of the nearest neighbors of instance $i$. In the experiments, we set the size of the spatial neighborhood $\mathcal{N}_i^{(spa)}$, and the size of the temporal neighborhood $\mathcal{N}_i^{(tem)}$ for each instance $i$ both to 2. $\lambda_1$, $\alpha$, and $\beta$ are the feature selection parameter, the spatial neighborhood parameter, and the temporal neighborhood parameter, respectively.

**TABLE 1: The RMSE on the testing set using Logistic Regression, $L_1$-regularized Logistic Regression, $L_1$-regularized Logistic Regression (ST), Neural Network, Autoencoder, ARIMA, RNN, and DAL, respectively.**

| | Root-Mean-Square Error (RMSE) | | | | |
|---|---|---|---|---|---|
| | Interpolation | Prediction | | | |
| | Hour 0 | Hour 1–12 | Hour 13–24 | Hour 25–36 | Hour 37–48 |
| Logistic Regression | 0.0671 | 0.0753 | 0.0907 | 0.0992 | 0.1103 |
| $L_1$-regularized Logistic Regression | 0.0680 | 0.0795 | 0.0935 | 0.1013 | 0.1141 |
| $L_1$-regularized Logistic Regression (ST) | 0.0675 | 0.0741 | 0.0884 | 0.0987 | 0.1089 |
| Neural Network | 0.0752 | 0.0823 | 0.0898 | 0.0943 | 0.1068 |
| Autoencoder | 0.0693 | 0.0714 | 0.0801 | 0.0894 | 0.0955 |
| Laplacian Regression | 0.0799 | 0.0800 | 0.0868 | 0.0932 | 0.0949 |
| ARIMA | 0.0714 | 0.0832 | 0.0913 | 0.1017 | 0.1150 |
| RNN | **0.0653** | 0.0679 | 0.0781 | 0.0837 | 0.0922 |
| DAL | **0.0667** | **0.0667** | **0.0724** | **0.0797** | **0.0877** |

**TABLE 2: Association analysis for different categories of features.**

| The Relevant Feature | $PM_{2.5}$, CO, $PM_{10}$, Wind Strength (North), Temperature, Wind Strength (East), $NO_2$, Precipitation, $SO_2$ |
|---|---|
| The Irrelevant Feature | Barometric Pressure, Humidity, $O_3$ |

## 4.2 Results and Discussions

For feature selection, we describe the number of features selected by DAL as a function of $\lambda_1$ in Figure 7(a), and the RMSE on the testing set using DAL as a function of $\lambda_1$ in Figure 7(b) when $\alpha = \beta = 0$. From Figure 7(a) we observe that the number of features selected by DAL decreases as the parameter $\lambda_1$ increases, which indicates the effectiveness of feature selection by the extra sparse layer. The mean activation of node $j$ in the extra sparse layer over the whole training set is defined as $MA_j = \frac{1}{m} \sum_{i=1}^{m} a_j^{(1)(i)}$. In the experiments, the range of $MA_j$ $(j = 1, ..., s_1)$ is in $[8.5906e^{-140}, 1]$ after the training when $\lambda_1 = 3$. The $j$th feature is removed from the network when $MA_j$ is very small. Here the feature selection can remove the least important features from the deep networks without decreasing the prediction accuracy a lot. The input features in our experiments are different air pollution data and meteorological data, which are all related to the prediction of the $PM_{2.5}$'s concentration. The feature selection keeps the major features, while removes the minor features, with the smallest decline of the predictor's performance, which reveals the importance of different input features of the black-box model. From Figure 7(b) we observe that, when $\lambda_1 = 2$, there are 133 features removed from the feature space, while the RMSE only increases 0.0013. To make a comparison, we remove 133 features randomly from the feature space 5 times, and train the model using the left features. The RMSE increases 0.0041 on average, which is much larger than that of the feature selection method, indicating that the extra sparse layer truly selects the most relevant features to the target during the training in the proposed model.

Note that the aim of the feature selection and analysis is not to increase the prediction accuracy, but to discover the importance of different input features to the predictions of the neural networks, reveal the main relevant factors to the variation of air quality, and provide a proof from data science to support the air pollution's prevention and control.

Most of the previous feature selection methods also decrease the prediction accuracy in many applications. However, our proposed model is flexible: when the online system needs high accuracy, the feature selection constraints can be removed; when one wants to know how the model makes predictions, adding the feature selection and analysis to the original prediction model reveals some inner mechanism of the black box.

In Table 1, we summarize the RMSE on the testing set using Logistic Regression, $L_1$-regularized Logistic Regression, $L_1$-regularized Logistic Regression (ST), Neural Network, Autoencoder, LapReg, ARIMA, RNN and DAL, respectively. For DAL in Table 1, $\lambda_1 = 2.5$, $\alpha = 10$, and $\beta = 15$. It is easy to understand that the RMSE on the testing set increases as the prediction hour increases for all the methods. From Table 1 we observe that DAL performs better than Logistic Regression, $L_1$-regularized Logistic Regression, $L_1$-regularized Logistic Regression (ST), Neural Network, Autoencoder, LapReg, and ARIMA both in the topics of interpolation (Hour 0) and prediction (Hour 1–48). RNN performs slightly better than DAL in the topic of interpolation, but much worse than DAL as the prediction hour increases in the topic of prediction. Because RNN only utilizes the temporal information of the data, and ignores the spacial information and the large amount of the unlabeled data. ARIMA also only utilizes the temporal information of the data, and it only uses the $PM_{2.5}$'s concentration as the input features, while ignores other related features. The performance of Logistic Regression is better than that of $L_1$-regularized Logistic Regression, for feature selection is embedded during the training process in $L_1$-regularized Logistic Regression, which may result in slight increase of the prediction error. Compared with Autoencoder, the extra sparse layer is added to DAL to achieve the purpose of feature selection. However, the performance of Autoencoder is inferior to the performance of DAL, for the usage of the spatio-temporal unlabeled data in DAL. Obviously, the characteristic of the spatio-temporal data, and the informa-tion of the unlabeled data are valuable to better exploit the geometric structure of the data. Embedding the spatio-temporal semi-supervised learning in the output layer of DAL is effective to decrease the learning error and improve the final performance. The performance of $L_1$-regularized Logistic Regression (ST) is better than that of $L_1$-regularized Logistic Regression, also indicating the effectiveness of the spatio-temporal constraints. In general, the deep model beats the regression model with the same spatio-temporal constraints for this application, because there exist a large amount of unlabeled data both in spatial dimension and temporal dimension, which can be utilized to pretrain the weights of the deep model. Laplacian Regression works in an inductive setting using both labeled and unlabeled training data and then test on new out-of-sample data. LapReg can tackle large-scale semi-supervised regression task. Compared with LapReg, our proposed model DAL can achieve the purpose of feature selection, with a much better performance of interpolation and prediction.

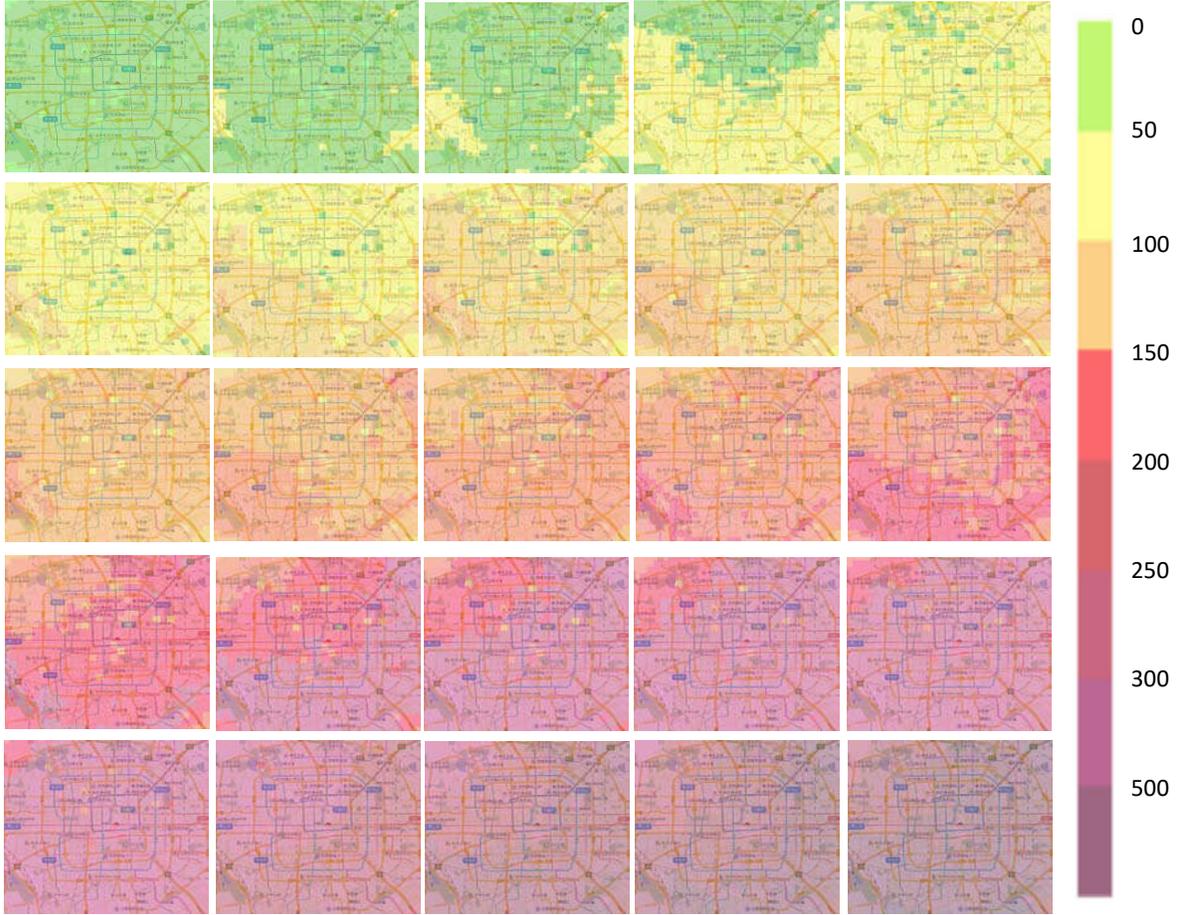

**Fig. 9:** The interpolation results of PM$_{2.5}$'s concentration using DAL to demonstrate a formation process of a haze happened in December in Beijing (within the Fifth Ring Road).

Despite of the moderate improvement of interpolation compared with the simple models, DAL is a unified model which makes prediction and interpolation synchronously without extra time consuming for interpolation. If we only deal with the interpolation task, there may be no need to use the deep learning based method to achieve the slight improvement. However, when we deal with the interpolation and prediction tasks simultaneously, adding 1-dim to the output layer of the deep networks can integrate interpolation with prediction, which almost does not increase the computation complexity of the prediction model.

To further explore the performance of the spatio-temporal semi-supervised learning, we describe the RMSE on the testing set using DAL as a function of $\alpha$ when $\lambda_1 = \beta = 0$ in Figure 8(a), and the RMSE on the testing set using DAL as a function of $\beta$ when $\lambda_1 = \alpha = 0$ in Figure 8(b). When $\alpha = 0$, it means there are no spatial neighborhood constraints; when $\beta = 0$, it means there are no temporal neighborhood constraints. From Figure 8 we observe that the RMSE decreases as $\alpha$ or $\beta$ increases, which shows that it is helpful to use the nearest neighbors from spatial neighborhood and temporal neighborhood to further improve the interpolation and prediction performance. The curve in Figure 8(b) declines more than the curve in Figure 8(a) does, indicating that the performance of temporal semi-supervised learning is better than that of spatial semi-supervised learning. For air quality related data, temporal correlation is more important than spatial correlation in the experiments.

Table 2 shows the result of association analysis in DAL for different categories of features when $\lambda_1 = 2.5$, $\alpha = 10$,

and $\beta = 15$. For simplicity, we treat each category of feature with different time stamps as a whole, and show the average results in Table 2. The most relevant features for the interpolation and prediction of $PM_{2.5}$'s concentration are: $PM_{2.5}$, CO, $PM_{10}$, Wind Strength (North), and Temperature. Because of the temporal correlation, the previous $PM_{2.5}$'s concentration is most relevant with the $PM_{2.5}$'s concentration in the following several hours. CO is a product of incomplete combustion for coal, oil, natural gas, etc. Since the data we used in the experiments are collected in winter of Beijing when the heating supply needs a great deal of coal, oil, and natural gas, it is reasonable that the CO's concentration is relevant with the $PM_{2.5}$'s concentration at this time of the year. $PM_{10}$ is relevant with $PM_{2.5}$ in most cases of the year based on our observations. Further, when there is a strong cold north wind, the haze disperses quickly. In winter, the haze always happens in the warm weather, and disperses in the cold weather consequently. Hence, the $PM_{2.5}$'s concentration is also relevant with Wind Strength (North) and Temperature. The most irrelevant features for $PM_{2.5}$'s concentration are: Barometric Pressure and Humidity, which is also reasonable for the Barometric Pressure and Humidity always change little in winter of Beijing, while $PM_{2.5}$'s concentration changes a lot.

Since the haze becomes a common weather phenomenon for Beijing, China during the whole winter these years, there must exit some general features that affect it. The feature analysis results show that our model is a reasonable and trustworthy model for the prediction of $PM_{2.5}$'s concentration, because they fit our intuitions on the formation of haze in winter of Beijing. And it truly reveals the most relevant factors based on the existing data we obtained, which is a proof from data science to support the air pollution's prevention and control. In the experiments, we also apply Algorithm 2 to the compared methods, Neural Network and Autoencoder. However, the results of association analysis in Neural Network and Autoencoder show no appreciable differences between different categories of features because there is no feature selection for these models. Combining with feature selection in DAL, association analysis reveals the main relevant and irrelevant features to the variation of the target. The proposed feature selection and analysis approach is a general method which does not limit to the feature set and the event used in the experiments, but can also be applied to various sets of features, different events, and many other applications such as medical diagnosis and terrorism detection.

Figure 9 shows the interpolation results of $PM_{2.5}$'s concentration using DAL to demonstrate a formation process of a haze happened in December in Beijing (within the Fifth Ring Road). From Figure 9 we observe that the haze starts from the south of the city, then diffuses to the north, indicating that the pollution sources locate in the south of Beijing. The results also fit our intuitions because there are lots of heavy industrial factories in Hebei province, south of Beijing.

In general, the experiments demonstrate that our proposed model DAL is an effective and promising solution to the topics of interpolation, prediction, and feature analysis of fine-gained air quality. The performance of DAL is superior to the performances of the compared methods for inter-polation and prediction. The main relevant and irrelevant features to the variation of $PM_{2.5}$'s concentration in Beijing are revealed by feature selection and association analysis in DAL. Further, the location of the pollution sources is also revealed by analysis of the interpolation results of $PM_{2.5}$'s concentration using DAL.

## 5 CONCLUSION

This paper studies three important topics in the area of urban air computing: the interpolation, prediction, and feature analysis of fine-gained air quality. The solutions to these topics can provide crucial information to support air pollution control, and consequently generate great societal and technical impacts. Most existing efforts focus on solving the three problems separately by establishing different models. In this paper, we develop a general and effective approach called DAL to unify the interpolation, prediction, feature selection and analysis of the fine-grained air quality into one model. In order to improve the performance of interpolation and prediction, we utilize the intrinsic characteristics of the spatio-temporal data and the information contained in the unlabeled data by embedding spatio-temporal semi-supervised learning on the output layer of neural network. We also propose a novel method to perform feature selection in the input layer of neural network, whose optimization is easy to solve and performance is well in removing the redundant or irrelevant features. Combining with feature selection, association analysis discovers the importance of different input features to the predictions of the neural networks. The proposed feature selection and analysis method has the ability to reveal some inner mechanism of the black-box deep network models. Extensive experiments are conducted on real data sources collected from Beijing, China, showing that DAL is superior to the comparison rivals when solving the topics of interpolation, prediction, and feature analysis of fine-gained air quality.

## ACKNOWLEDGMENTS

This work was mainly performed at the time that Zhongang Qi had a close research collaboration with Prof. Xi Li as a researcher at NEC Laboratories China. This work was supported by the grant from NEC Laboratories China, in part by the National Basic Research Program of China under Grant 2015CB352302, and in part by the National Natural Science Foundation of China under Grant U1509206, Grant 61472353, Grant 61751209, and Grant 62572041.

## REFERENCES

[1] R. Tibshirani, "Regression shrinkage and selection via the lasso," *Journal of the Royal Statistical Society. Series B (Methodological)*, vol. 58, no. 1, pp. 267–288, 1996.

[2] M. Yuan and Y. Lin, "Model selection and estimation in regression with grouped variables," *Journal of the Royal Statistical Society: Series B (Statistical Methodology)*, vol. 68, pp. 49–67, 2006.

[3] L. Li, X. Zhang, J. Holt, J. Tian, and R. Piltner, "Spatiotemporal interpolation methods for air pollution exposure," in *Symposium on Abstraction, Reformulation, and Approximation*, 2011.

[4] Y. Zheng, F. Liu, and H.-P. Hsieh, "U-air: When urban air quality inference meets big data," in *Proceedings of the 19th ACM SIGKDD International Conference on Knowledge Discovery and Data Mining*, ser. KDD '13, 2013, pp. 1436–1444.

[5] H.-P. Hsieh, S.-D. Lin, and Y. Zheng, "Inferring air quality for station location recommendation based on urban big data," in *Proceedings of the 21th ACM SIGKDD International Conference on Knowledge Discovery and Data Mining*, ser. KDD '15, 2015, pp. 437–446.

[6] M. Dong, D. Yang, Y. Kuang, D. He, S. Erdal, and D. Kenski, "PM$_{2.5}$ concentration prediction using hidden semi-markov model-based times series data mining," *Expert Syst. Appl.*, vol. 36, no. 5, pp. 9046–9055, Jul. 2009.

[7] S. Thomas and R. B. Jacko, "Model for forecasting expressway pm2.5 concentration – application of regression and neural network models." *Journal of the Air & Waste Management Association*, vol. 57, no. 4, pp. 480–488, 2007.

[8] Y. Zheng, X. Yi, M. Li, R. Li, Z. Shan, E. Chang, and T. Li, "Forecasting fine-grained air quality based on big data," in *Proceedings of the 21th ACM SIGKDD International Conference on Knowledge Discovery and Data Mining*, ser. KDD '15, 2015.

[9] X. Zhou, W. Huang, N. Zhang, W. Hu, S. Du, G. Song, and K. Xie, "Probabilistic dynamic causal model for temporal data," in *Neural Networks (IJCNN), 2015 International Joint Conference on*, July 2015, pp. 1–8.

[10] K. P. Singh, S. Gupta, and P. Rai, "Identifying pollution sources and predicting urban air quality using ensemble learning methods," *Atmospheric Environment*, vol. 80, pp. 426 – 437, 2013.

[11] C. Rosenberg, M. Hebert, and H. Schneiderman, "Semi-supervised self-training of object detection models," in *Seventh IEEE Workshop on Applications of Computer Vision*, 2005.

[12] A. Blum and T. Mitchell, "Combining labeled and unlabeled data with co-training," in *Proceedings of the Eleventh Annual Conference on Computational Learning Theory*, ser. COLT' 98, 1998, pp. 92–100.

[13] B. Maeireizo, D. Litman, and R. Hwa, "Co-training for predicting emotions with spoken dialogue data," in *Proceedings of the ACL 2004 on Interactive Poster and Demonstration Sessions*, ser. ACLdemo '04. Association for Computational Linguistics, 2004.

[14] Y. Li, Z. Qi, Z. M. Zhang, and M. Yang, "Learning with limited and noisy tagging," in *Proceedings of the 21st ACM International Conference on Multimedia*, ser. MM '13, 2013, pp. 957–966.

[15] R. Socher, J. Pennington, E. H. Huang, A. Y. Ng, and C. D. Manning, "Semi-supervised recursive autoencoders for predicting sentiment distributions," in *Proceedings of the Conference on Empirical Methods in Natural Language Processing*. Association for Computational Linguistics, 2011, pp. 151–161.

[16] J. Weston, F. Ratle, and R. Collobert, "Deep learning via semi-supervised embedding," in *the 25th International Conference on Machine Learning*, 2008.

[17] N. Cressie and C. K. Wikle, *Statistics for Spatio-Temporal Data*. Wiley, 2011.

[18] Y. Saeys, I. Inza, and P. Larranaga, "A review of feature selection techniques in bioinformatics," *Bioinformatics*, vol. 23, no. 19, pp. 2507–2517, 2007.

[19] R. Setiono and H. Liu, "Neural-network feature selector," *Neural Networks, IEEE Transactions on*, vol. 8, no. 3, pp. 654–662, 1997.

[20] R. Tibshirani, "The lasso method for variable selection in the cox model," *Statistics in Medicine*, vol. 16, pp. 385–395, 1997.

[21] J. Lokhorst, *The lasso and generalised linear models*. Honors Project. University of Adelaide, Adelaide, 1999.

[22] V. Roth, "The generalized lasso," *IEEE Transactions on Neural Networks*, vol. 15, pp. 16–28, 2004.

[23] B. Krishnapuram, L. Carin, M. Figueiredo, and A. Hartemink, "Sparse multinomial logistic regression: fast algorithms and generalization bounds," *IEEE Transactions on Pattern Analysis and Machine Intelligence*, vol. 27, pp. 957–968, 2005.

[24] A. Y. Ng, "Feature selection, l1 vs. l2 regularization, and rotational invariance," in *Proceedings of the Twenty-first International Conference on Machine Learning*, ser. ICML '04, 2004.

[25] S. Bakin, *Adaptive regression and model selection in data mining problems*. PhD Thesis. Australian National University, Canberra., 1999.

[26] L. Meier, S. V. D. Geer, and P. Bhlmann, "The group lasso for logistic regression," *Journal of the Royal Statistical Society: Series B (Statistical Methodology)*, vol. 70, pp. 53–71, 2008.

[27] B. Efron, T. Hastie, I. Johnstone, R. Tibshirani *et al.*, "Least angle regression," *The Annals of statistics*, vol. 32, no. 2, pp. 407–499, 2004.

[28] X. Chen, Q. Lin, S. Kim, J. G. Carbonell, E. P. Xing *et al.*, "Smoothing proximal gradient method for general structured sparse regression," *The Annals of Applied Statistics*, vol. 6, no. 2, pp. 719–752, 2012.

[29] T. T. Wu and K. Lange, "Coordinate descent algorithms for lasso penalized regression," *The Annals of Applied Statistics*, pp. 224–244, 2008.

[30] A. Ng, "Sparse autoencoder," *CS294A Lecture Notes*, 2011.

[31] Y. Chauvin and D. E. Rumelhart, *Backpropagation: theory, architectures, and applications*. Psychology Press, 1995.

[32] Y. Bengio, "Learning deep architectures for ai," *Foundations and Trends in Machine Learning*, vol. 2, no. 1, pp. 1–127, 2009.

[33] R. en Fan, K. wei Chang, C. jui Hsieh, X. rui Wang, and C. jen Lin, "Liblinear: A library for large linear classification," *Journal of Machine Learning Research*, vol. 9, pp. 1871–1874, 2008. [Online]. Available: http://jmlr.org/papers/volume9/fan08a/fan08a.pdf

[34] V. Sindhwani, P. Niyogi, M. Belkin, and S. Keerthi, "Linear manifold regularization for large scale semi-supervised learning," in *Proc. of the 22st ICML Workshop on Learning with Partially Classified Training Data*, 2005.

[35] J. Wu and S. Wei, *Time series analysis*. Hunan Science and Technology Press, ChangSha, 1989.

[36] T. Mikolov, M. Karafiát, L. Burget, J. Cernocký, and S. Khudanpur, "Recurrent neural network based language model." in *Interspeech*, vol. 2, 2010, p. 3.